\documentclass{ieeetj}
\usepackage{cite}
\usepackage{amsmath,graphicx,amsfonts,amsthm,amssymb}
\usepackage{algorithmic}
\usepackage{graphicx,color}
\usepackage{textcomp}
\usepackage{xcolor}
\usepackage{hyperref}
\hypersetup{hidelinks=true}
\usepackage{algorithm,algorithmic}
\def\BibTeX{{\rm B\kern-.05em{\sc i\kern-.025em b}\kern-.08em
    T\kern-.1667em\lower.7ex\hbox{E}\kern-.125emX}}
\AtBeginDocument{\definecolor{tmlcncolor}{cmyk}{0.93,0.59,0.15,0.02}\definecolor{NavyBlue}{RGB}{0,86,125}}

\usepackage{bbold}
\usepackage{hyperref}
\usepackage[capitalise]{cleveref}
\usepackage{wrapfig}
\usepackage{microtype}
\usepackage{xfrac}
\usepackage{lipsum}
\usepackage{float}
\usepackage{stfloats}
\usepackage{orcidlink}
\usepackage{mathtools}
\usepackage[group-separator={,}]{siunitx}
\usepackage{subcaption}

\newtheorem{lemma}{Lemma}

\newcommand{\GG}{\mathcal{G}}
\newcommand{\DD}{\mathcal{D}}
\newcommand{\Sc}{\mathcal{S}}
\newcommand{\MM}{\mathcal{M}}
\newcommand{\Aa}{\mathbf{A}}
\newcommand{\Bb}{\mathbf{B}}
\DeclareMathOperator{\trace}{tr}
\DeclareMathOperator{\MSE}{MSE}
\DeclareMathOperator{\Pa}{Pa}
\newcommand{\V}[1]{{\boldsymbol{\mathbf{#1}}}}

\def\OJlogo{\vspace{-4pt}}
\def\seclogo{\vspace{10pt}}

\def\authorrefmark#1{\ensuremath{^{\textbf{#1}}}}

\begin{document}
\receiveddate{6 September, 2023}
\reviseddate{23 December, 2023}
\accepteddate{26 December, 2023}
\publisheddate{}
\currentdate{2 January, 2024}
\doiinfo{}

\markboth{\textsc{Dagma-DCE}}{Waxman {et al.}}

\title{\textsc{Dagma-DCE}: Interpretable, Non-Parametric Differentiable Causal Discovery}

\author{Daniel Waxman \orcidlink{0009-0004-0168-5547}\authorrefmark{1}, Graduate Student Member, IEEE, Kurt Butler \orcidlink{0000-0002-1520-4909}\authorrefmark{1}, Graduate Student Member, IEEE, and Petar M. Djuri\'c \orcidlink{0000-0001-7791-3199}\authorrefmark{1}, Life Fellow, IEEE}
\affil{Department of Electrical and Computer Engineering, Stony Brook University, Stony Brook, NY 11794 USA}
\corresp{Corresponding author: Petar M. Djuri\'c (\texttt{petar.djuric@stonybrook.edu})}
\authornote{This work was supported by the National Science Foundation under Awards 2212506 and 2021002.}

\begin{abstract}
    We introduce \textsc{Dagma-DCE}, an interpretable and model-agnostic scheme for differentiable causal discovery. Current non- or over-parametric methods in differentiable causal discovery use opaque proxies of ``independence'' to justify the inclusion or exclusion of a causal relationship. We show theoretically and empirically that these proxies may be arbitrarily different than the actual causal strength. Juxtaposed to existing differentiable causal discovery algorithms, \textsc{Dagma-DCE} uses an interpretable measure of causal strength to define weighted adjacency matrices. In a number of simulated datasets, we show our method achieves state-of-the-art level performance. We additionally show that \textsc{Dagma-DCE} allows for principled thresholding and sparsity penalties by domain-experts. The code for our method is available open-source at \url{https://github.com/DanWaxman/DAGMA-DCE}, and can easily be adapted to arbitrary differentiable models.
\end{abstract}

\begin{IEEEkeywords}
    Causal discovery, interpretable machine learning, differential causal effect
\end{IEEEkeywords}

\maketitle

\section{INTRODUCTION}
The discovery of causal relationships is the subject of many scientific questions, in medicine, economics, the social sciences, and more, and is therefore a question of utmost importance in science and engineering. Given an observational signal, one may therefore be interested in learning the causal structure underlying the data. Under Pearl's formalism of causality \cite{pearl2009causality}, the causal structure is captured by a \emph{causal graph}, which is a directed acyclic graph (DAG) with edges denoting direct causation of one variable to another. 

A core principle of causal inference is the notion of \emph{intervention}, where one can interact with the system that produces signals. This is in contrast to causal inference from \emph{observational} data, where we can only observe the outputs of a system, as is typically the case in signal processing or machine learning. Unfortunately, causal discovery using observational data is generally ill-posed and requires strong assumptions about the system we observe \cite[pp. 135]{peters2017elements}. Different methods may use different assumptions, where the validity of the assumptions may be extremely domain- or problem-specific. Even when assumptions are made to ensure (at least partial) identifiability of the causal graph, causal discovery can be extremely expensive, in terms of both the computation and the amount of required data.

One major class of methods, referred to as \emph{constraint-based} methods, is based on the observation that a causal graph encodes a set of conditional independencies (CIs). By assuming that all CIs are encoded in the graph, an assumption known as \emph{causal faithfulness}, constraint-based methods utilize statistical CI tests to determine which causal graphs are consistent with the data.

While some of the earliest successful causal discovery algorithms are constrained-based, such as the PC Algorithm and Fast Causal Inference \cite{spirtes2000causation}, they can be very costly. Namely, they are both expensive computationally, scaling exponentially with the number of variables in the worst case, and they are expensive in terms of data. Additionally, CI tests tend not to be very powerful, and constrained-based causal discovery methods require conditional data for many different values and combinations of random variables. Moreover, causal faithfulness is not a testable assumption in general, though some weaker versions are \cite{zhang2008detection}.

An alternative approach to causal discovery using observational data is to use \emph{score-based} methods. These methods attempt to find a causal model $\MM$ and its associated causal graph $\GG$, which maximizes a score function $\Sc(\MM, \DD)$ for observational data $\DD = \{ \V{x}_1, \V{x}_2, \dots, \V{x}_N \}$ \cite[pp. 148]{peters2017elements}. For example, one can search for the causal model which maximizes the likelihood or minimizes the mean square error (MSE). In practice, one typically optimizes a version of these objectives which is penalized for model complexity, such as the Bayesian information criterion. For certain classes of causal models, this approach is simple and effective --- indeed, early attempts at score-based causal discovery were made contemporaneously to early constraint-based approaches \cite{chickering2002optimal}. Despite their simplicity, score-based methods suffer from the super-exponential search space of DAGs with $d$ variables, as well as identifiability issues \cite[pp. 149]{peters2017elements}.

Fundamentally troubling to score-based methods is the difficulty of optimization in discrete spaces, which tend to be combinatorially more difficult than continuous objectives. To address this, \textsc{Notears} (short for \emph{\textbf{N}on-combinatorial \textbf{O}ptimization via \textbf{T}race \textbf{E}xponential and \textbf{A}ugmented lag\textbf{R}angian for \textbf{S}tructure learning}) \cite{zheng2018dags} proposes a novel algebraic constraint on adjacency matrices, such that the constraint is satisfied if and only if the adjacency matrix $\Aa$ represents a DAG. \textsc{Notears} then performs constrained continuous optimization on a penalized MSE objective to learn a causal graph without explicitly exploring the discrete space of all DAGs.

\textsc{Notears} %
was originally defined for linear models, with the adjacency matrix entries being coefficients of the corresponding linear functions. \textsc{Notears+} \cite{zheng2020learning} extends \textsc{Notears} to nonlinear functions and achieves state-of-the-art performance on simulated data. In lieu of linear coefficients, \textsc{Notears+} uses opaque proxies to define $\Aa$, providing advantages in optimization but at the cost of interpretability and generality of the results. 

Several methods have provided alternative acyclicity constraints which are faster to compute or are otherwise preferable in optimization. One striking example is \textsc{Dagma} (short for \emph{\textbf{DAG}s via $\mathbf{M}$-matrices with \textbf{A}cyclicity}) \cite{bello2022dagma}, which in practice can optimize up to an order of magnitude faster than \textsc{Notears}. 

Our contribution is to provide a novel learning objective that results in obtaining interpretable weighted causal graphs. Our method has several important strengths. It is model-agnostic in nature, with the ability to use any twice-differentiable model trainable with gradient descent. Furthermore, the sparsity penalty used in the scoring function $\Sc(\cdot, \cdot)$ is also model-agnostic. Lastly, we can interpret the causal graphs with the differentiable causal effect (DCE) metric \cite{butler2022differential}. We base our optimization on the method of \textsc{Dagma}, leading us to name our method \textsc{Dagma-DCE}.

The rest of this paper is structured as follows: in Section II, we review score-based causal discovery and the current state-of-the-art in differentiable causal discovery. In Section III, we show that the adjacency matrix returned by \textsc{Notears} and \textsc{Dagma} is not interpretable, first by a theoretical argument and then by a toy example. We propose a new graph definition and optimization problem in Section IV, forming \textsc{Dagma-DCE}, and discuss its interpretability in Section V. Finally, we run experiments on synthetic data in Section VI showing that \textsc{Dagma-DCE} performs competitively with other methods and returns legitimately different adjacency matrices than \textsc{Dagma}. Concluding remarks are provided in Section VII.

\section{BACKGROUND}
We begin by reviewing some concepts of causal inference, namely structural causal models (SCMs), which formalize causality. We then introduce differentiable causal discovery methods, which utilize continuous optimization for score-based causal discovery.

\subsection{Structural Causal Models and Causal Discovery}
SCMs provide a formalism for causality, where causal relationships are described by a DAG and accompanying functional assignments. To make this more clear, let $\V{x} = [x_1, \dots, x_d]$ be $d$-dimensional random vector, and $\GG$ be a DAG whose vertices are the components of $\V x$. Then, according to $\GG$, an SCM describes each component $x_j$ as a function of its parents $\Pa(x_j)$ and independent noise $z_j$, $f_j(\Pa(x_j), z_j)$, i.e.,
\begin{equation} \label{eq:scm_defn}
    x_j \coloneqq f_j\left(\Pa(x_j), z_j\right), \quad j = 1, \dots, d.
\end{equation}
We use the notation $\coloneqq$ to reinforce the somewhat philosophical point that these are \emph{assignments}, and not merely equivalence \cite{peters2017elements}. We will frequently refer to the SCM of a system as simply the ``causal model'' $\MM$.

The goal of structure learning is then to identify the SCM of a system, given data $\mathcal{D}$. If one cannot interact with the system, then we must make assumptions on the form of each $f_j$ and the distribution of the noise $z_j$ to uniquely determine the SCM \cite[pp. 139]{peters2017elements}. These assumptions can be parametric, for example, assuming each $f_j$ is linear and each $z_j$ is non-Gaussian \cite{shimizu2006linear}, or they can be non-parametric, for example, assuming the $f_j$ are strictly nonlinear with the $z_j$ being additive and Gaussian \cite[Corollary 31]{peters2014causal}. Once we make such a restriction to our model, we call it \emph{identifiable}.

After establishing identifiability, there are several families of techniques for learning the SCM from observational data. Here, we focus on score-based methods, though the interested reader is pointed to \cite{vowels2022d} for a recent, comprehensive review. In score-based causal discovery, we find a causal model $\MM$ which maximizes a score function $\Sc(\MM, \DD)$. In practice, we parameterize models by some vector ${\V{\theta}}$ and then find the model $\MM_{\V{\theta}}$ which optimizes $\Sc(\cdot, \cdot)$:

\begin{equation} \label{eq:score_based_defn}
    \hat{{\V{\theta}}} = \arg \max_{\V{\theta}} \Sc(\MM_{\V{\theta}}, \DD) \textrm{ s.t. } \GG_{\V{\theta}} \text{ is a DAG.}
\end{equation}

The main drawback of na\"ive score-based methods is that they have to search over the space of DAGs with $d$ nodes, which grows extremely quickly. For $d=5$, this number is already \num{29281}; for $d=14$, it is already greater than $10^{36}$ \cite[A003024]{oeis}. Therefore, for even small graphs, one must resort to greedy or approximate techniques \cite[pp. 150]{peters2017elements}.

\subsection{Differentiable Causal Discovery}
Differentiable causal discovery methods are score-based methods that recast \cref{eq:score_based_defn} as a constrained continuous optimization problem; as before, let $\Sc(\cdot, \cdot)$ be a score function that reflects how well a model $\MM_{\V{\theta}}$ explains the data $\DD$, and $\Aa_{\V{\theta}} \in \mathbb{R}_{\geq 0}^{d \times d}$ be a weighted adjacency matrix derived from $\MM_{\V{\theta}}$. Then, using a function $h(\cdot)$, which is $0$ if and only if $\Aa_{\V{\theta}}$ represents a DAG, the optimization problem \cref{eq:score_based_defn} becomes
\begin{equation} \label{eq:general_problem}
    \begin{aligned}
    \max_{{\V{\theta}}\in{\V{\theta}}} \quad & \Sc(\MM_{\V{\theta}}, \DD), \\
    \textrm{s.t.} \quad & h(\Aa_{\V{\theta}}) = 0.
    \end{aligned}
\end{equation}

The earliest differentiable causal discovery algorithm with a DAG constraint is \textsc{Notears} \cite{zheng2018dags}, which uses a trace of a nonlinear function on the adjacency matrix $\Aa_{\V{\theta}}$, i.e.,
\begin{equation*}
    h_\textsc{Notears}(\Aa_{\V{\theta}}) = \trace\left(\exp(\Aa_{\V{\theta}} \odot \Aa_{\V{\theta}})\right) - d,
\end{equation*}
where $\odot$ is the Hadamard (elementwise) product. The motivation for this constraint is that powers of the adjacency matrix $\Aa_{\V{\theta}}^k$ reflect how many closed walks of length $k$ are present in the corresponding graph $\GG_{\V{\theta}}$. Since DAGs have no closed walks of any length, $h(\Aa_{\V{\theta}})$ is $0$ precisely when $\Aa_{\V{\theta}}$ is a DAG.

Quickly, several simplifications on the same constraint were derived. For example, one need not calculate the entire matrix exponential, with a polynomial sufficing instead \cite{wei2020dags}. 

The alternative constraint that we use is \textsc{Dagma} \cite{bello2022dagma}; with $s > 0$, the \textsc{Dagma} constraint function is
\begin{equation} \label{eq:dagma_constraint}
    h_\textsc{Dagma}(\Aa_{\V{\theta}}) = - \log\det\left(s\mathbb{1}_d - \Aa_{\V{\theta}} \odot \Aa_{\V{\theta}}\right) + d\log s,
\end{equation}
where $\mathbb{1}_d$ is the $d \times d$ identity matrix. This constraint is motivated by the fact that $\Aa_{\V{\theta}}$ describes a DAG if and only if it is nilpotent. While the constraint given by \cref{eq:dagma_constraint} is, in general, a relaxation of nilpotency, the authors of \textsc{Dagma} characterize the set of matrices for which zeroness of \cref{eq:dagma_constraint} is equivalent to acyclicity, called $\mathbf{M}$-matrices. They then show that the set of $\mathbf{M}$-matrices includes all DAGs, and that the gradient of $h_\textsc{Dagma}(\cdot)$ is sufficiently well-behaved, to enable constrained optimization within $\mathbf{M}$-matrices.

It remains to specify the model $\MM_{\V{\theta}}$, the score function $\Sc$, and how to derive $\GG_{\V{\theta}}$ from $\MM_{\V{\theta}}$. The approach introduced in \textsc{Notears+} and used in \textsc{Dagma}\footnote{In general, \textsc{Dagma} refers to both linear and nonlinear models using \cref{eq:dagma_constraint}. To avoid confusion, \textsc{Dagma} exclusively refers to the nonlinear case here.} is to define
\begin{equation} \label{eq:adjacency_def}
    {{\Aa_{\V{\theta}}}_{ij}} \triangleq \lVert \partial_i f_j \rVert_{L^2} = \sqrt{\int_{\mathbb{R}^d} \left(\partial_i f_j(\V{x})\right)^2 \, d\V{x}},
\end{equation}
where $f_j$ is the function $\V{x} \mapsto x_j$ according to $\MM_{\V{\theta}}$. To guarantee \cref{eq:adjacency_def} is well-defined, it is required that $\lVert \partial_i f_j \rVert_{L^2}$ is finite. Accordingly, the \textsc{Notears+} authors restrict $\MM_{\V{\theta}}$ to a space of functions called the Sobolev space $H^1(\mathbb{R}^d)$.

However, the \textsc{Notears+} authors argue that for causal discovery, the zeroness of ${{\Aa_{\V{\theta}}}_{ij}}$ is what matters, and so a condition equivalent to $ {{\Aa_{\V{\theta}}}_{ij}} = 0 $ is derived for several classes of models. For example, in a multilayer perception (MLP) network, the $L^2$ norm of the first layer in the neural network is chosen. The score function is defined to be a sparsity-penalized MSE\footnote{\textsc{Dagma} uses the log-likelihood instead for benchmarks, but we use the MSE to simplify comparisons, and avoid assumptions on the data-generating process.}, 
\begin{equation} \label{eq:score_function}
    \Sc(\MM_{\V{\theta}}, \DD) = -(\MSE(\MM_{\V{\theta}}, \DD) + \lambda_1 g(\MM_{\V{\theta}})),
\end{equation}
where $\lambda_1$ is a hyperparameter and $g(\MM_{\V{\theta}})$ defines a sparsity penalty. For MLPs, \textsc{Notears+} and \textsc{Dagma} use the $\ell_1$ norm of the first layer MLP weights.

In practice, it is very difficult to exactly follow the constraint in \cref{eq:general_problem}. For this reason, as well as to avoid false positives, an additional thresholding step is performed, where $\Aa_{ij}$ is set to $0$ if it is less than some predetermined hyperparameter $C$.

\section{NON-INTERPRETABILITY OF \textsc{Dagma}}
A key motivation for \textsc{Dagma-DCE} is that existing methods utilize a non-interpretable weighted adjacency matrix $\Aa_{\V{\theta}}$. In this section, we argue theoretically that the adjacency matrix of \textsc{Dagma} is not interpretable and subsequently show empirically that \textsc{Dagma} fails to return intuitively-interpretable results on linear SCMs. 

For the purposes of this section, we restrict ourselves to models where $f_j$ is modeled with an MLP network, though our arguments can be adjusted for other cases. The theoretical section can safely be skipped, but shows explicitly that adjacency matrices derived from the first layer of a neural network may be quite different than those defined by derivatives.

\subsection{Theoretical Argument}

For the purposes of this section, let $f_j \colon \mathbb{R}^d \to \mathbb{R}$ be an MLP with weight matrices $\Aa^{(1)}, \dots, \Aa^{(M)}$ and activation function $\sigma(\cdot)$. Then the $(M-1)$-hidden layer MLP $f_j$ can be expressed as 
\begin{equation} \label{eq:mlp_defn}
    f_j(\V{x}) = \Aa^{(M)} \sigma\left(\Aa^{(M-1)} \sigma\left( \cdots \sigma\left(\Aa^{(1)} \V{x}\right)\right)\right).
\end{equation}

Despite the exact analytic result that 
\begin{equation} \label{eq:equivalence_of_norms}
    \lVert \partial_i f_j \rVert_{L^2} = 0 \iff \lVert \Aa^{(1)}_{i\cdot} \rVert_{L^2} = 0,
\end{equation}
differentiable causal discovery methods rely on thresholding small, nonzero values. The following lemma shows that when using sigmoid activation functions, as is done in \textsc{Notears+} and \textsc{Dagma}, the two norms in \cref{eq:equivalence_of_norms} may be arbitrarily different.

\begin{lemma}
    Let $\sigma(\cdot)$ denote the sigmoid activation function. Then for any $\delta, \epsilon> 0$, there exists an MLP $f_j$ with weight matrices $\Aa^{(1)}, \dots, \Aa^{(M)}$ such that $\lVert \Aa^{(1)}_{i\cdot} \rVert_{L^2} < \epsilon$ but $\lVert \partial_i f_j \rVert_{L^2}> \delta$.
\end{lemma}
\begin{IEEEproof}
    As \eqref{eq:mlp_defn} is a repeated composition of smooth functions, it suffices to show the result for the $1$-hidden layer MLP
    \begin{equation*}
        g(\V{x}) = \Aa^{(2)} \sigma\left(\Aa^{(1)} \V{x}\right).
    \end{equation*}
    If $\Aa^{(2)}$ is a $K \times H$ matrix, then
    \begin{equation*}
        [g(x)]_k = \sum_h^H \Aa^{(2)}_{kh} \sigma\left( \sum_{j=1}^d \Aa^{(1)}_{hj} x_j \right).
    \end{equation*}
    For notational clarity, let 
    \begin{equation*}
        z_h = \sigma\left( \sum_{j=1}^d \Aa^{(1)}_{hj} x_j \right).
    \end{equation*}
    Additionally, note that $\sigma'(\cdot) > 0$. Then so long as $\Aa^{(1)}_{hi}$ is nonzero for some $h$, we have by chain rule that 
    \begin{equation*}
        \left\lvert\frac{\partial z_h}{\partial x_i}\right\rvert > \epsilon'
    \end{equation*}
    for some $\epsilon' > 0$. Then by directly calculating the norm,
    \begin{align*}
        \lVert \partial_i f_j \rVert_{L^2}^2 &= \int \sum_{k=1}^K \left(\frac{\partial g_k}{\partial x_i}\right)^2 \, dx_i \\
        &= \int \sum_{k=1}^K \left(\sum_{h=1}^H \Aa^{(2)}_{kh} \frac{\partial z_h}{\partial x_i} \right)^2 \, dx_i\\
        &\geq \int \sum_{k=1}^K \left(\sum_{h=1}^H \Aa^{(2)}_{kh} \epsilon' \right)^2 \, dx_i.
    \end{align*}
    we conclude that it is unbounded. Therefore, we can choose $\Aa^{(2)}$ such that the function norm is arbitrarily large, and in particular, larger than $\delta$.
\end{IEEEproof}

While this result may not be particularly ``generic,'' in the sense that the resulting function $f_j$ may be unlikely to be learned in practice, it does show that $\partial_i f_j$ can be arbitrarily different than any nonzero $\Aa^{(1)}$. This situation only gets worse for the more popular rectified linear unit (ReLU) activation function, where $\lVert \partial_i f_j \rVert_{L^2}$ is arbitrary for a \emph{fixed} $f_j$. This is formalized in the following lemma:

\begin{lemma}
    Let $\sigma(\cdot)$ denote the ReLU activation function. Then for any $s>0$ and $\Aa^{(1)},\dots, \Aa^{(M)}$, there exist matrices $\Bb^{(1)}, \dots, \Bb^{(M)}$ such that 
    \begin{align*}
        f_j(\V{x}) &= \Aa^{(M)} \sigma\left(\Aa^{(M-1)} \sigma\left( \cdots \sigma\left(\Aa^{(1)} \V{x}\right)\right)\right)\\
        &= \Bb^{(M)} \sigma\left(\Bb^{(M-1)} \sigma\left( \cdots \sigma\left(\Bb^{(1)} \V{x}\right)\right)\right),
    \end{align*}
    and $\lVert \Bb^{(1)}_{i\cdot} \rVert_{L^2} = s$.
\end{lemma}
\begin{IEEEproof}
    Note that for any real $z$, the ReLU function is invariant to positive rescaling: $\sigma(z) = \sigma(az)/a$ for any $a > 0$. This implies, more generally, that $\mathbf{S}^{-1} \Aa^{(2)} \sigma(\mathbf{S} \Aa^{(1)} \V{x}) = \Aa^{(2)} \sigma(\Aa^{(1)} \V{x})$, where $\mathbf{S}$ is a diagonal matrix \cite{bona2021parameter}. The lemma then immediately follows by setting $\Bb^{(1)} = \mathbf{S} \Aa^{(1)}$ and $\Bb^{(2)} = \mathbf{S}^{-1} \Aa^{(2)}$, with the appropriate choice of $\mathbf{S}$.
\end{IEEEproof}

\subsection{Empirical Result with a Linear SCM} 
\label{sec:dagma_linear_scm}
\begin{figure*}[htp]
     \centering
     \begin{subfigure}[b]{0.48\textwidth}
        \centering
        \includegraphics[width=2.55in]{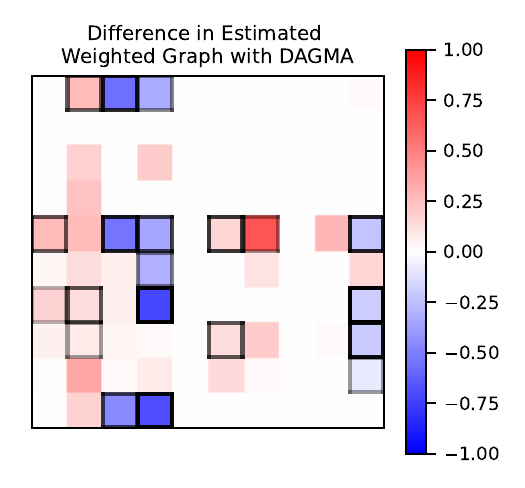}   
        \vspace{-1.5em}
        \caption{\textsc{Dagma}}
        \label{fig:LinearDifferencesDAGMA} 
     \end{subfigure}
     \hfill
     \begin{subfigure}[b]{0.48\textwidth}
        \centering
        \includegraphics[width=2.55in]{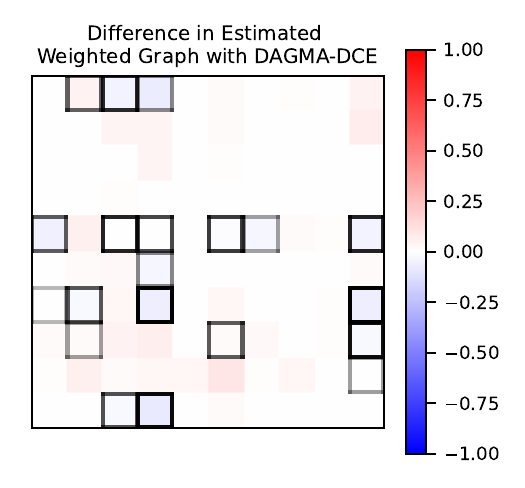}
        \vspace{-1.5em}
        \caption{\textsc{Dagma-DCE}}
        \label{fig:LinearDifferencesDCE}                 
     \end{subfigure}
        \caption{The difference between the magnitude of the true derivatives in a linear causal model to the magnitude of the weighted graph in \textsc{Dagma} for a random $10 \times 10$ Erd\"os-R\'enyi directed graph with $20$ expected edges. Gray boxes surrounding each cell denote the magnitude of the ground-truth linear coefficient.}
        \label{fig:LinearDifferences}
\end{figure*}
In the lemmas above, we have shown that the adjacency matrix of \textsc{Notears+} and \textsc{Dagma} may be arbitrarily different than the $L^2$ function norm. Here, we illustrate this fact in practice by applying \textsc{Dagma} to a linear structural equation model (SEM), and showing that the returned adjacency matrix does not correspond to the intuitive notion of causal strength.

For this purpose, we simulated an Erd\"os-R\'enyi (ER) graph with $10$ random variables and an expectation of $20$ edges, creating a DAG as in \cite{zheng2018dags}. Given a DAG, a  SEM assigns functional relationships between a variable and its parents. In this case, a linear SEM is assigned by randomly sampling linear coefficients from $\mathcal{U}\left((-2.0, -0.5) \cup (0.5, 2.0)\right)$. 

The use of a linear SEM allows for an obvious ground-truth weighted graph to compare to, namely the linear coefficients. We then ran \textsc{Dagma} with each $f_j$ being modeled as an MLP with one hidden layer of $10$ units and compared the resulting weighted adjacency matrix to the ground truth in \cref{fig:LinearDifferencesDAGMA}. For the purposes of this demonstration, we tested with $\lambda_1 \in \{0, 10^{-4}, 10^{-3}, 10^{-2}, 10^{-1}\}$ and chose the value which minimized the Frobenius norm $\lVert \Aa_\text{est} - \Aa_\text{true} \rVert_F$ to give a favorable comparison. This resulted in using $\lambda_1 = 10^{-2}$, with all other hyperparameters set to their default values.

Two things are visually apparent: first, \textsc{Dagma} learns the graph topology quite well, including one erroneous edge in an otherwise perfect prediction. Second, however, is that the weights recovered by \textsc{Dagma} do not obviously correspond to the interpretable, linear coefficients, i.e., it cannot easily be interpreted as a ``strength of causation.'' Even worse, it is not easy to predict when \textsc{Dagma} is over- or under-estimating the strength of an edge. While we defer a definition of \textsc{Dagma-DCE} and discussion of \cref{fig:LinearDifferencesDCE} to the subsequent sections, a view of \cref{fig:LinearDifferencesDCE} shows a much closer resemblance to linear coefficients. It is important to note that this does not indicate ``errors'' in \textsc{Dagma}, but instead that the magnitude of the weighted graph \emph{is not defined} to be the true ``causal strength'' --- their very definition has lost meaning compared to the linear case.

\section{PROPOSED SOLUTION}
In this section, we provide a definition of the learning objective for \textsc{Dagma-DCE}, and subsequently discuss practical details of implementing such a scheme. In particular, we discuss how \textsc{Dagma-DCE} is model-agnostic in both formulation and implementation, and we also explore simple methods for accelerating and pre-training models.

\subsection{Definition of the Optimization Problem}
An initial solution would be to estimate \cref{eq:adjacency_def} directly; however, we argue $H^1(\mathbb{R}^d)$ is the wrong space to define models over. Not only are many interesting functions outside of $H^1(\mathbb{R}^d)$, but also simple functions, for example linear functions, do not have a finite Sobolev norm over all of $\mathbb{R}^d$. Additionally, from a data-centric perspective, there is little reason to care about the behavior of the model where there is a vanishingly small probability mass. 

Instead, we choose to define the adjacency matrix as the $L^2$ norm of the derivative $\partial_i f_j$ with respect to the probability distribution of $\V{X}$, which we denote $\mathbb{P}^\V{X}$. While the theory of Sobolev spaces plays no further role in our work, to adapt the language of \cite{zheng2020learning}, models are defined over the \emph{weighted Sobolev space $H^1(\mathbb{R}^d, \mathbb{P}^\V{X})$} \cite{Kufner1984}. This is simply to say that the models' function and first (weak) derivatives are square-integrable with respect to $\mathbb{P}^\V{X}$. We then offer the alternative definition and Monte Carlo approximation
\begin{align} \label{eq:dagma_dce_defn}
        \Aa_{ij} \triangleq \lVert \partial_i f_j \rVert_{L^2(\mathbb{P}^\V{X})} &= \sqrt{\int_{\mathbb{R}^d} \left(\partial_i f_j(\V{x}) \right)^2 p_{\V{X}}(\V{x}) \, d\V{x}} \notag \\
        & \approx \sqrt{\frac{1}{N}\sum_{n=1}^N \left(\partial_i f_j(\V{x}_n) \right)^2},
\end{align}
where $p_\V{X}(\V x)$ is the density of $\mathbb{P}^\V{X}$ with respect to the Lebesgue measure, and $\V{x}_n \sim p_\V{X}(\V x)$. By noting that the entries of $\Aa$ are the root-mean-square entries of the Jacobian matrix, it is then straightforward to calculate $\Aa_{ij}$ for any differentiable model, with the above being a Monte Carlo approximation over empirical Jacobian matrices.

This definition of $\Aa$ also allows for a principled sparsity penalty --- indeed, if $\partial_i f_j$ is square-integrable with respect to $\mathbb{P}^\V{X}$, then Jensen's inequality implies that it is also integrable, which allows us to define and approximate
\begin{equation} \label{eq:dagma_dce_l1}
    g(\MM_{\V{\theta}}) = \lVert \partial_i f_j \rVert_{L^1(\mathbb{P}^\V{X})} \approx \frac{1}{N} \sum_{n=1}^N \lvert \partial_i f_j(\V{x}_n) \rvert.
\end{equation}

Combining \cref{eq:general_problem,eq:dagma_constraint,eq:score_function,eq:dagma_dce_defn,eq:dagma_dce_l1}, we arrive at the optimization problem
\begin{equation} \label{eq:dagma_dce_problem}
    \begin{aligned}
    \min_{{\V{\theta}}\in{\V{\theta}}} \quad & \MSE(\MM_{\V{\theta}}, \DD) + \lambda_1 \lVert \partial_i f_j \rVert_{L^1(\mathbb{P}^\V{X})} \\
    \textrm{s.t.} \quad & -\log\det\left(s\mathbb{1}_d - \Aa_{\V{\theta}} \odot \Aa_{\V{\theta}}\right) + d\log s = 0.
    \end{aligned}
\end{equation}
Similar to \textsc{Dagma}, we use a central path optimization technique to solve the constrained optimization problem. 

Our approach has the same convergence properties as \textsc{Notears} and \textsc{Dagma}, using similar assumptions. Namely, we assume causal sufficiency, i.e., a lack of latent confounders, and identifiability, which ensures that the optima of the score are the true causal graph. There are many options for ensuring identifiability, for example, thrice-differentiable strictly-nonlinear additive noise models suffice \cite{peters2014causal}. Additionally, while it is not necessary for the optimization problem to be well-defined, the interpretability of \textsc{Dagma-DCE} relies on once-differentiability of the underlying $f_j$. Namely, this excludes discrete or mixed-type data, since how a model interpolates functions between discrete labels is arbitrary.

\subsection{\textsc{Dagma-DCE} is Model Agnostic}
The derivatives in \cref{eq:dagma_dce_defn} may be available analytically for many models. For the simplest case, take $\MM_{\V{\theta}}$ to be linear, i.e.,
\begin{equation*}
    x_j \coloneqq \sum_{i=1}^d \beta_{ij} x_i.
\end{equation*}
Then the linear coefficients $\beta_{ij}$ are the derivatives used in \cref{eq:dagma_dce_defn}. Note that \cref{eq:dagma_dce_problem} then becomes the problem of linear \textsc{Dagma}, meaning \textsc{Dagma-DCE} is a legitimate generalization.

Simple (semi-)analytic expressions are also available for additive models:
\begin{equation} \label{eq:additive_model_defn}
    x_j \coloneqq \sum_{k=1}^K \beta_{jk} g_k(\V{x}),
\end{equation}
in which case 
\begin{equation*}
    \partial_i f_j = \sum_{k=1}^K \beta_{jk} \frac{\partial g_k}{\partial x_i}.
\end{equation*}

Analytic expressions of coefficients are even available for several classes of non-parametric models. For example, by the representer theorem \cite[pp. 132]{williams2006gaussian}, a Gaussian process (GP) posterior can be written in the form of \cref{eq:additive_model_defn}, where $g(\V{x})$ depends on the kernel $\kappa(\cdot, \cdot)$ and $\beta_{j1}, \dots, \beta_{jn}$ are easily calculated:
\begin{equation}
      \begin{split}
    f_j &\sim \mathcal{GP}(\V{m}, \kappa); \\
    x_j &\coloneqq f_j(\V{x}),
  \end{split}
\;\Longleftrightarrow\;
  \begin{split}
    x_j \coloneqq \sum_{n=1}^N \beta_n \kappa(\V{x}, \V{x}_n)
  \end{split}.
\end{equation}
Closed-form expressions for several common kernels, including the automatic relevance detection squared exponential (ARD-SE) and Mat\'ern-3/2 kernels, are derived in \cite{butler2022differential}.

Furthermore, even when analytic expressions become cumbersome to derive by hand, the derivative can often be obtained via automatic differentiation (AD) tools. Among many other differentiable models, a notable case is with MLPs. It is also the case for extensions of \textsc{Notears+}-like objectives to time series, for example the one-dimensional convolutional neural networks used in \textsc{NTS-Notears} \cite{sun2023nts}.

\subsection{Practical Considerations} \label{sec:practical_considerations}
Several general comments are made in Section~4.1 of \cite{bello2022dagma} which also hold for \textsc{Dagma-DCE}, particularly concerning the optimization details. For example, the hyperparameter $s$ can be changed freely, even between optimization iterations. This can affect performance, as $s$ controls the volume of the feasible region. The authors of \cite{bello2022dagma} recommend starting with $s=1$ and slowly reducing its value. Additionally, different optimizers can be used to solve sub-problems along the central path; \textsc{Dagma} uses Adam \cite{Kingma15}, which we also use. Finally, the use of central path methods also creates additional hyperparameters, including central path coefficients and weight initializations, which are discussed.

While backward passes over the Jacobian is indeed expensive and a drawback of the \textsc{Dagma-DCE}, the Monte Carlo approximation can be effectively parallelized with respect to $N$, particularly on GPUs. Therefore, while scaling with $d$ is inherently difficult, scaling with $N$ is quite inexpensive.

In practice, we additionally see benefits from pre-training for \textsc{Dagma-DCE}, namely by training a single iteration of \textsc{Dagma}. Intuitively, while \textsc{Dagma} and \textsc{Dagma-DCE} are fundamentally different optimization problems, their underlying task is the same. Therefore, a single iteration of \textsc{Dagma} serves as a good starting point, and saves costly backward passes over the Jacobian.

Finally, \textsc{Dagma-DCE} has several hyperparameters, including the strength of $L_1$ regularization $\lambda_1$ and thresholds. In the next section, we discuss how the formulation of \textsc{Dagma-DCE} makes the threshold hyperparameter interpretable. However, the issue of setting $\lambda_1$ remains. One option, explored in \cite{zheng2020learning} and \cite{sun2023nts}, is to optimize $\lambda_1$ with respect to some metric using cross-validation if labeled data/causal graph pairs exist. How to choose $\lambda_1$ when these labeled pairs are not available is an interesting task for future research.

\section{INTERPRETABILITY OF \textsc{Dagma-DCE}}
The interpretability of the \textsc{Dagma-DCE} approach stems from the observation that the partial derivatives, $\partial_i f_j$, can be interpreted through the lens of causal strength. We begin this section with general notes on causal strengths, followed by the differential causal effect, which forms the basis of interpretability for \textsc{Dagma-DCE}. We revisit the empirical example of Section~III-\ref{sec:dagma_linear_scm} and show that \textsc{Dagma-DCE} recovers the intuitive result. Finally, we discuss how thresholding in \textsc{Dagma-DCE} is a fundamentally interpretable operation, allowing the incorporation of prior knowledge from experts and decision-makers.

\subsection{Measuring Causal Strength}
The quantification of causal strength is a highly nontrivial task, with several competing definitions \cite{janzing2013quantifying}. 
Despite this, we find it helpful to ground our reasoning in the linear case, where an unambiguous notion of causal strength is available. In particular, the linear model coefficients quantify exactly how a change in one variable will induce changes in its children. Consider a linear SCM with coefficients given by a matrix $[\Bb]_{ij} = \beta_{ij}$ and additive noise $z_j$:
$$
x_j := \sum_{i=1}^d \beta_{ij} x_i + z_j, \qquad j=1,...,d.
$$
Then the entry $\beta_{ij}$ represents the direct influence of $x_i$ to $x_j$. Note that ${\Bb}$ can be viewed as a causal graph, provided that it is acyclic, or equivalently, nilpotent.

Occasionally, these direct coefficients summarize all causal information that $x_i$ holds about $x_j$. Indeed, if $\Bb$ is viewed as a graph, this is the case if the only path $x_i$ and $x_j$ is the direct one, in which case $\Bb_{ij}$ is the total causal strength. Otherwise, in general, we must sum across all paths in the graph, and multiply the coefficients along each path. This can be expressed compactly by a geometric sum,
$$
\mathbf{T} = \sum_{\ell=1}^d \Bb^\ell.
$$
Causal sufficiency of the system then tells us that the entries $\mathbf{T}_{ij}$ of the matrix $\mathbf{T}$ are sufficient to describe how any change to $x_i$ will affect $x_j$, and thus, $\mathbf{T}$ can be useful to ask counterfactual questions or to understand the impact of various causes to a single effect.

\subsection{Differential Causal Effect and \textsc{Dagma-DCE}}
The DCE provides a nonlinear version of the linear model causal strength by considering the local behavior of the model \cite{butler2022differential}. The Jacobian matrix, $\mathcal{J}_f$, replaces the role of ${\Bb}$, describing how changes to a parent affect the child node when there are no other paths in the system. Similarly, the geometric sum of the Jacobian also defines the total causal effect from source to effect, for small changes. 

Relating to \textsc{Dagma-DCE}, the adjacency matrix defined in \eqref{eq:dagma_dce_defn} can be interpreted as the root-mean-squared DCE --- as a result, $\mathbf{A}_{ij}\neq0$ if and only if the DCE, $\partial_i f_j$, is not identically zero. The magnitude of $\mathbf{A}_{ij}$ quantifies how strong the interaction is, in terms of the energy of the DCE functions. %

\subsection{Empirical Results with a Linear SCM}

We quickly revisit the experiment of Section~III-\ref{sec:dagma_linear_scm}, where \textsc{Dagma} was fit to a linear SCM, and its weighted adjacency matrix compared to the ground truth. The setup is nearly identical, with the same parameterization of each function $f_j$. The main difference is that we do not tune the sparsity penalty hyperparameter, instead we simply set $\lambda_1 = 0$. Like \textsc{Dagma}, the \textsc{Dagma-DCE} method obtains an accurate estimate of the graph topology, recovering the underlying graph perfectly. Unlike \textsc{Dagma}, we additionally recover the ``intuitive'' notion of causal strength. 

\subsection{Thresholding in \textsc{Dagma-DCE}}
An ad-hoc component of \textsc{Notears+}-like methodologies is the thresholding step. Indeed, across different datasets, the thresholding hyperparameter can have large consequences to accuracy, as shown in \cite{zheng2020learning}. However, many conceivable causal inference tasks will not have any validation set to calibrate the thresholding parameter with, meaning the user will have to make a heuristic estimate.

On the contrary, \textsc{Dagma-DCE}'s weighted adjacency matrix entries are interpretable, so the choice of a threshold carries direct meaning to the modeler or expert. This decision can even be made on an edge-by-edge basis, or across (a possibly rescaled version of) $\Aa$. 

For example, one may decide to normalize the entry $A_{ij}$ of by the marginal variance of $x_j$ for the purposes of thresholding, reminiscent of ANOVA \cite{sober1988apportioning}. An alternative scheme could be normalizing by the sum of the $j$th column; while subadditivity of norms means this does not actually scale by the total derivative $\partial f_j$, it can be viewed as scaling by the total causal effect. Finally, since the entries of $\Aa$ are now unconstrained, one could attempt thresholding methods developed for Granger causality, for example via clustering \cite{plub2019estimation} or denoising \cite{bayati2023granger}. How this thresholding step should be interpreted and executed forms an interesting question for future work.

\begin{figure*}[tp]
     \centering
    \begin{subfigure}[b]{0.48\textwidth}
        \centering
        \includegraphics[width=3.25in]{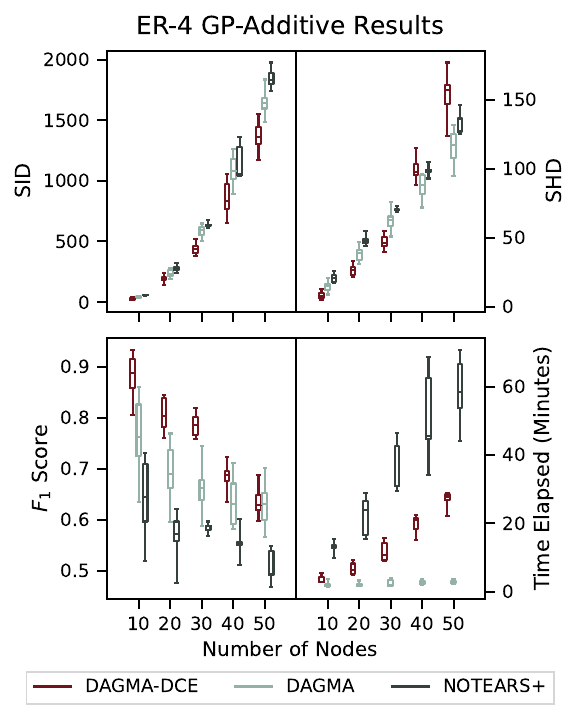}
        \caption{}\label{fig:GP_Add_Results} 
     \end{subfigure}
     \hfill
     \begin{subfigure}[b]{0.48\textwidth}
        \centering
        \includegraphics[width=3.25in]{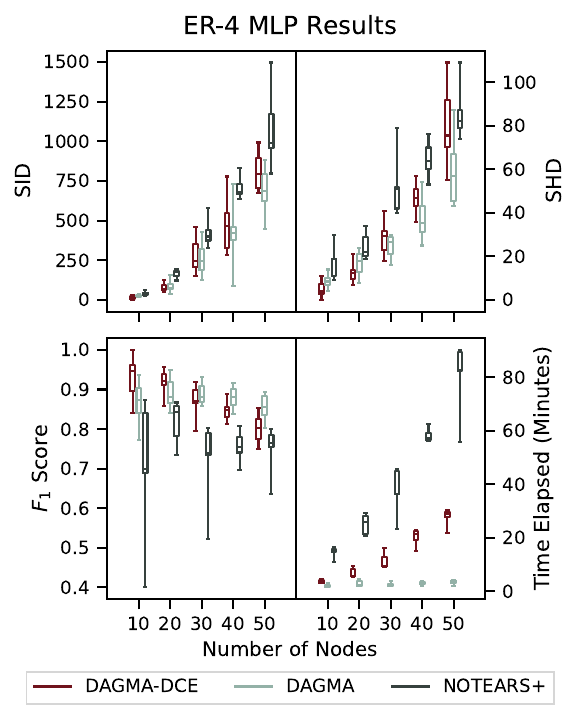}
        \caption{}\label{fig:MLP_Results}
     \end{subfigure} 

    \caption{Resulting SID (top left), SHD (top right), $F_1$ Score (bottom left), and time elapsed (bottom right) for random data generated from \protect{(\subref{fig:GP_Add_Results})} the ER-4 GP-additive model and \protect{(\subref{fig:MLP_Results})} the ER-4 MLP model, as detailed in \cref{sec:synthetic_data}. Boxes show the median and quartiles across $T=10$ trials for \textsc{Dagma} and \textsc{Dagma-DCE}, and $T=5$ trials for \textsc{Notears}, with whiskers showing the minimum and maximum values.}
    \label{fig:Results}
\end{figure*}

\section{EXPERIMENTS}\label{sec:experiment}

We perform several experiments similar to those conducted by the authors of \textsc{Dagma} \cite{bello2022dagma}, providing comparisons to \textsc{Dagma} and \textsc{Notears+}. We begin with a brief discussion of metrics used to evaluate results, followed by experimental details, results, and discussion.

\subsection{Metrics}
Evaluating results in causal discovery is complex in the sense that several different metrics exist with different goals and calculations. The two most basic metrics we report are the structural Hamming distance (SHD), which calculates the number of incorrect edges a graph contains, and the $F_1$ score, which is the geometric mean of precision and recall. 

Additionally, we include the structural intervention distance (SID) \cite{peters2015structural}, which is arguably more well-suited to causal inference. The SID is motivated by the idea that, when calculating causal effects via ``adjustment sets,'' some errors are more severe than others. For example, including spurious relations does not affect causal effects, while reversing the order of an edge does. The SID calculates the number of edges $i \to j$ such that the adjustment set corresponding to the calculation of the causal effect of $x_i$ on $x_j$ is not valid.

Finally, we report the time elapsed in minutes in each of our synthetic benchmarks. 

\subsection{Experimental Details} \label{sec:synthetic_data}
We replicate a common synthetic dataset in the causal discovery literature, generating ER-$4$ DAGs according to the procedure of \cite{zheng2020learning}, with functional relationships assigned in one of two ways. The first way is using additive GPs: 
\begin{equation*}
    f_j(\V{x}) = \sum_{i \in \Pa(x_j)} g_{ij}(x_i),
\end{equation*}
where each $g_{ij}$ is sampled from an RBF GP with lengthscale $1$. 
The other is with a random MLP network with hidden size $100$ and a sigmoid activation. Crucially, all weights are samples from $\mathcal{U}\left((-2.0, -0.5) \cup (0.5, 2.0)\right)$. 

As noted in Section~IV-\ref{sec:practical_considerations}, the hyperparameters of each method may generally be tuned with cross-validation. Nonetheless, to facilitate a fair comparison between methods, we use a set of ``defaults'' as in \cite{zheng2020learning,bello2022dagma,sun2023nts}. For \textsc{Dagma} and \textsc{Notears+}, we use the hyperparameters noted in Appendix C.2.2 of \cite{bello2022dagma}. However, since the sparsity penalty of \textsc{Dagma-DCE} is fundamentally different than that of other methods, it is difficult to find a ``default'' setting for a fair comparison. We chose $\lambda_1 = 3.5 \times 10^{-2}$, which provided a sparsity penalty of similar magnitude in the MLP experiments. This is imperfect, but allows for the initial score functions of \textsc{Dagma} and \textsc{Dagma-DCE} to be similar without unfairly optimizing $\lambda_1$. Similarly, we used $0.25$ as thresholds for comparison, which provides a similar true positive rate in the MLP experiments. Additionally, due to the added cost of \textsc{Dagma-DCE}, we set the maximum number of iterations to $10\%$ of the corresponding values in \textsc{Dagma}. 

All experiments were run on a NVIDIA Titan RTX GPU. The authors' implementations of \textsc{Notears+}\footnote{\url{https://github.com/xunzheng/notears}} and \textsc{Dagma}\footnote{\url{https://github.com/kevinsbello/dagma}} were used, with slight modifications to allow GPU acceleration. All methods (including \textsc{Dagma-DCE}) used the same MLP architecture used in the experiments of \cite{zheng2020learning,bello2022dagma}.

\subsection{Results \& Discussion}
Our results show that \textsc{Dagma-DCE} achieves the state-of-the-art on common synthetic datasets: in additive GPs \protect{(\cref{fig:GP_Add_Results})}, the median SID is universally lower than \textsc{Dagma} or \textsc{Notears+}. Similarly, the $F_1$ score is higher for all but $D=50$ nodes, and the SHD is significantly lower for $d \leq 30$ nodes. While \textsc{Dagma-DCE} is indeed more computationally expensive than \textsc{Dagma}, it is still substantially faster than \textsc{Notears} across all graph sizes.

In the MLP experiments \protect{(\cref{fig:MLP_Results})}, our results are somewhat less impressive, but still competitive. This is despite what we argue is a generating process that gives a distinct advantage to \textsc{Dagma} and \textsc{Notears+}: by recalling that these methods perform thresholding on the norm of layer weights, sampling weights from $\mathcal{U}\left((-2.0, -0.5) \cup (0.5, 2.0)\right)$ builds-in identifiability in a perhaps unrealistic manner.

Finally, without asserting what the ``correct'' magnitude of weighted adjacency matrices is, it is interesting to note that \textsc{Dagma} and \textsc{Dagma-DCE} return adjacency matrices in which the relative magnitudes differ considerably. We measure this with Kendall's $\tau$-b and the Spearman correlation coefficient, which measure rank correlation. These both take on values in $[-1, 1]$, where a value of $\pm 1$ indicates complete agreement or disagreement, and a value of $0$ indicates that orderings are unrelated. The resulting values for our synthetic experiments can be found in \cref{tab:correlation_table}.

\setlength\intextsep{0pt}
\begin{table}[h]
\centering
\caption{Rank correlations between adjacency matrix entries of \textsc{Dagma} and \textsc{Dagma-DCE} across all trials for synthetic datasets, given as the mean plus-minus one standard deviation.} \label{tab:correlation_table}
\begin{tabular}{l|l|l}
\hline
Dataset      & Kendall's $\tau\text{-b}$ & Spearman's $\rho$ \\ \hline
Additive GPs & $0.40 \pm 0.09$           & $0.53 \pm 0.11$   \\ 
MLP          & $0.55 \pm 0.06$           & $0.74 \pm 0.07$   \\ \hline
\end{tabular}
\end{table}

\section{CONCLUSION}
We reformulated the objectives of \textsc{Notears+} and \textsc{Dagma} in order to extract a principled, interpretable weighted adjacency matrix. With this comes principled, interpretable notions of an $\ell_1$ sparsity penalty, and thresholding the resulting graph.

A drawback of this method is its computational complexity, requiring a backward pass over the Jacobian; however, empirical results indicate it is still much faster than some other competing methods that use an augmented Lagrangian scheme for optimization. Furthermore, we were able to achieve state-of-the-art performance on synthetic datasets with small-to-medium-sized graphs. Interesting directions for future work include applications to real datasets and a formal workflow for determining sparsity and thresholding hyperparameters.

\bibliographystyle{IEEEbib}
\bibliography{references}

\vfill\pagebreak

\end{document}